\documentclass[runningheads]{llncs}
\usepackage{graphicx}
\usepackage{amsmath,amssymb} 
\usepackage{color}
\usepackage[width=122mm,left=12mm,paperwidth=146mm,height=193mm,top=12mm,paperheight=217mm]{geometry}
\usepackage[]{hyperref}
\usepackage[nocompress]{cite}
\usepackage{xspace}
\usepackage{tabularx}
\usepackage{booktabs}
\usepackage{multirow,rotating}
\usepackage{etoolbox}
\usepackage{subcaption}
\usepackage{epstopdf}
\usepackage{siunitx}
\usepackage{chngpage}

\definecolor{darkblue}{rgb}{0,0.1,0.5}
\hypersetup{colorlinks,
            linkcolor=darkblue,
            anchorcolor=darkblue,
            citecolor=darkblue}

\newcommand{\eq}[1]{Eq. \ref{eq:#1}}

\newcommand{\Eg}{\emph{E.g.,}\xspace}
\newcommand{\eg}{\emph{e.g.,}\xspace}
\newcommand{\etal}{\emph{et al.}\xspace}
\newcommand{\ie}{\emph{i.e.,}\xspace}

\begin{document}
\pagestyle{headings}
\mainmatter

\title{VideoLSTM Convolves, Attends and Flows\\ for Action Recognition} 
\titlerunning{VideoLSTM Convolves, Attends and Flows for Action Recognition}
\authorrunning{Z. Li, E. Gavves, M. Jain, and C.G.M. Snoek}

\author{Zhenyang Li, Efstratios Gavves, Mihir Jain, and Cees G. M. Snoek}
\institute{QUVA Lab, University of Amsterdam}

\maketitle

\begin{abstract} 
We present a new architecture for end-to-end sequence learning of actions in video, we call \emph{VideoLSTM}. Rather than adapting the video to the peculiarities of established recurrent or convolutional architectures, we adapt the architecture to fit the requirements of the video medium. Starting from the soft-Attention LSTM, VideoLSTM makes three novel contributions. First, video has a spatial layout. To exploit the spatial correlation we hardwire convolutions in the soft-Attention LSTM architecture. Second, motion not only informs us about the action content, but also guides better the attention towards the relevant spatio-temporal locations. We introduce motion-based attention. And finally, we demonstrate how the attention from VideoLSTM can be used for action localization by relying on just the action class label. Experiments and comparisons on challenging datasets for action classification and localization support our claims.
\end{abstract}

%
\section{Introduction} 
Is a video a stack of ordered images? Or is it a stack of short-term motion patterns encoded by the optical flow?
When an action appears in an image, what is the spatio-temporal extent of that action to the semantics and should it be encoded as a memory?
All these questions highlight various video properties that need to be addressed when modelling actions, \eg what is the right appearance and motion representation, how to transform spatio-temporal video content into a memory vector, how to model the spatio-temporal locality of an action?

Answering these questions and modelling \emph{simultaneously all} these video properties is hard, as the visual content varies dramatically from video to video and from action to action. For actions like \emph{shaving} or \emph{playing the piano} it is the context that is indicative. For actions like \emph{juggling} and \emph{running} the motion is defining. And, for actions like \emph{tiding up} or \emph{dribbling with the football} one would need to combine appearance (\eg the soccer field), the motion (\eg running with the ball) as well the relative temporal transitions from previous frames (\eg passing through players) to arrive at the right prediction. Given the sheer complexity, the literature has shown preference to pragmatic approaches that excel in specific aspects, \eg emphasizing mainly on the video's local appearance~\cite{KarpathyCVPR14,SimonyanNIPS14}, local motion~\cite{jain2013wflow, wang2013iccv, SimonyanNIPS14} and dynamics~\cite{Fernando2015a} or local spatio-temporal patterns~\cite{JiPAMI13,TranICCV15}. Ideally, however, the video should be treated as a separate medium. Namely the video model should attempt to address most, if not all of the video properties.

A class of models that seems like an excellent initial candidate for modelling the video to its fullest are the Long Short-Term Memory (LSTM) networks. LSTM, originally proposed in~\cite{HochreiterNC97}, has recently become popular~\cite{DonahueCVPR15,NgCVPR15,SrivastavaICML15,SharmaICLR16} due to its sequence modelling capabilities.
LSTM is a recurrent, end-to-end neural architecture, which on one hand may receive several and different inputs, \eg activations from deep networks processing raw RGB or optical flow pixels, while on the other hand may preserve internally a memory of what has happened over a long time period. What is more, it was recently shown~\cite{XuICML15, SharmaICLR16} that LSTMs can be reinforced so that they focus more on specific input dimensions. This soft-Attention concept is an excellent fit to the action locality in videos. It should be no surprise, therefore, that several works~\cite{DonahueCVPR15,NgCVPR15,SrivastavaICML15,SharmaICLR16} have proposed to use \emph{out-of-the-box} LSTM networks to model videos and action classification.

\begin{figure}[t!]
    \centering
    \includegraphics[width=0.9\linewidth]{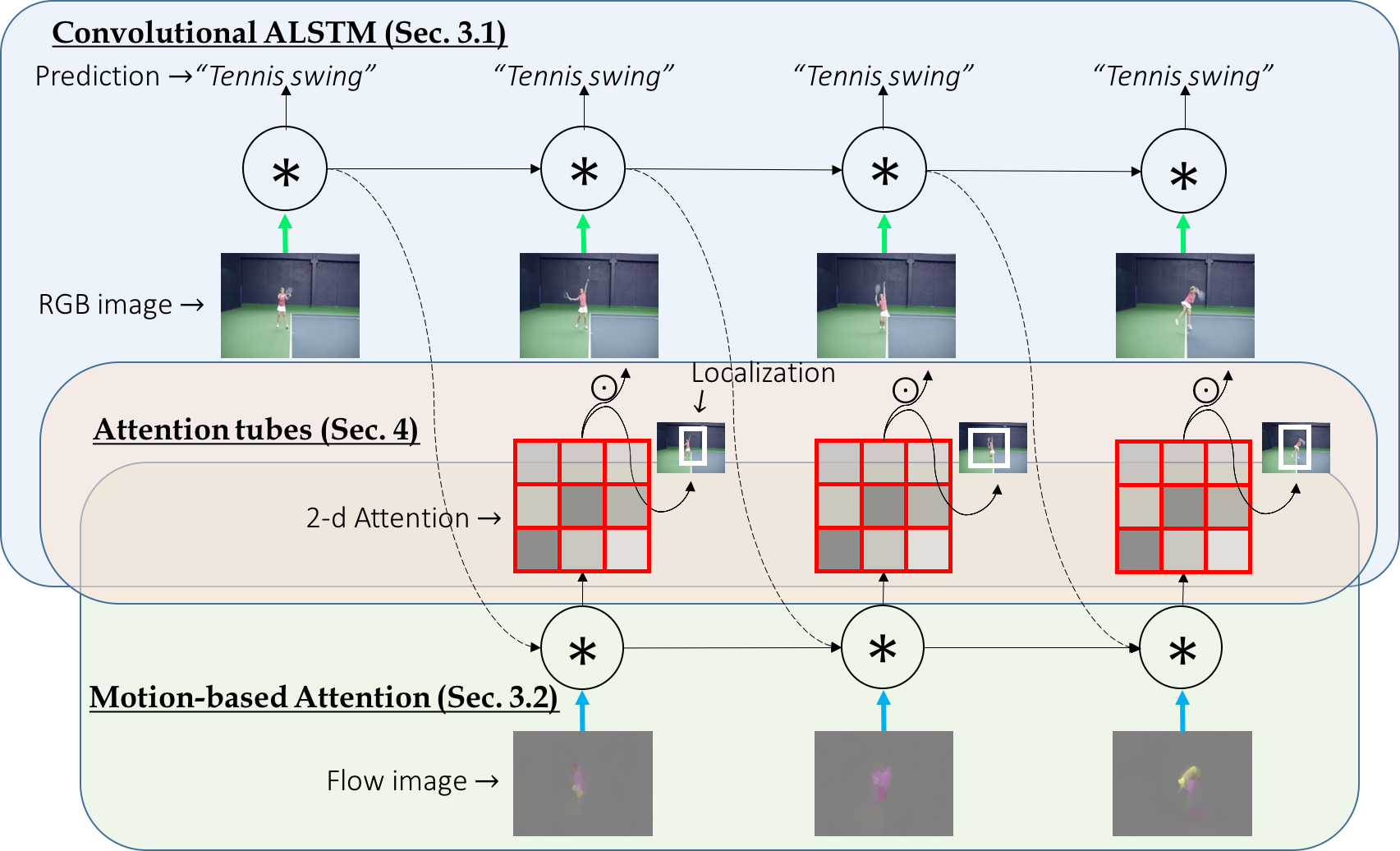}
    \caption{\textbf{The proposed \emph{VideoLSTM} network.} The blue container stands for the \emph{Convolutional ALSTM} (Sec.~\ref{sec:conv-lstm}), the green container stands for the motion-based attention (Sec.~\ref{sec:motion-based-attention}), while in the pale red container we rely on attention maps for action localization (Sec.~\ref{sec:localization}).}
    \label{fig:structure}
\end{figure}

Although at first glance LSTMs seem ideal, they face certain shortcomings. Both the standard LSTM \cite{HochreiterNC97}  as well as the Attention LSTM treat all incoming data as vectors, even if they have a spatial structure. \Eg a spatial input of \texttt{conv5} activations would still be vectorized, before fed to the LSTM. Ignoring the spatial correlation between locally neighboring frame pixels, however, goes against the very nature of the video medium.
The models that are known to model best such spatial correlations are convolutional neural networks \cite{lecun1998gradient}.
Their secret is in the convolutions, which are differentiable, ensure better shift, scale and distortion invariance by leveraging local receptive fields, share weights, and pool information locally, especially when stacked one after the other in very deep architectures~\cite{szegedy2015going,SimonyanVeryDeep,HeResNet15}.
Adding convolutions to LSTM was very recently shown to be effective~\cite{shi2015nips} for radar map forecasting.
In modelling videos, however, our experiments show that convolutions alone do not suffice and attention must also be considered.
Hence, to reckon the spatio-temporal nature of the video when using an LSTM, we must hardwire the LSTM network with convolutions and spatio-temporal attentions.

In this work we advocate and experimentally verify that to model videos accurately, we must adapt the LSTM architecture to fit the video medium, not vice versa. Such a model must address all video properties \emph{simultaneously} and not few in isolation.
In fact, our experiments reveal that modelling a subset only of the properties brings little, if any, improvement, while a joint treatment results in consistent improvements. We propose \textit{VideoLSTM}, a new recurrent neural network architecture intended for action recognition, particularly.

VideoLSTM makes three novel contributions that attempt to address the aforementioned video properties in a joint fashion.
First, VideoLSTM recognizes that video frames have a spatial layout.
Hence the video frame encoding as well as the attention should be spatial too.
We introduce convolutions to exploit the spatial correlations in images.
Second, VideoLSTM recognizes that the motion not only informs us about the action content, but also guides better the attention towards the relevant spatio-temporal locations.
We propose a motion-based attention mechanism that relies on a shallow convolutional neural network, instead of the traditional multi-layer perceptron of Attention-LSTM models.
And finally, we demonstrate that by only relying on video-level action class labels the attention from VideoLSTM can be used for competitive action localization without being explicitly instructed to do so. 
Before detailing our architecture (summarized in Fig.~\ref{fig:structure}), we first provide a synopsis of related work.

%
\section{Related work}
The literature on action recognition in video is vast and too broad for us to cover here completely. We reckon and value the impact of traditional video representations, \eg \cite{wang2013iccv,jain2013wflow,SadanandCVPR2012,peng2014bag,pengECCV14stackedFisher,Fernando2015a,Lan2015}, and recent mixtures of shallow and deep encodings \eg \cite{WangCVPR15,JainCVPR15}. Here we focus on deep end-to-end alternatives that have recently become popular and powerful. 

\textbf{ConvNet architectures.} 
One of the first attempts of using a deep learning architecture for action recognition is by Ji \etal~\cite{JiPAMI13}, who propose 3-d convolutional neural networks. A 3-d convolutional network is the natural extension of a standard 2-d convolutional network to cover the temporal domain as a third dimension. However, processing video frames directly significantly increases the learning complexity, as the filters need to model both appearance and motion variations. To compensate for the increased parameter complexity larger datasets are required. Indeed, Tran \etal~\cite{TranICCV15} recently demonstrated that 3-d convolutional networks trained on massive sport video datasets~\cite{KarpathyCVPR14} yield significantly better accuracies.

To avoid having to deal with this added complexity of spatio-temporal convolutional filters, Simonyan and Zisserman~\cite{SimonyanNIPS14} proposed a two-stream architecture to learn 2-d filters for the optical flow and appearance variations independently. In order to capture longer temporal patterns, several frames are added as multiple consecutive channels as input to the network. To account for the lack of training data a multi-task setting is proposed, where the same network is optimized for two datasets simultaneously. Similar to~\cite{SimonyanNIPS14} we also use optical flow to learn motion filters. Different from~\cite{SimonyanNIPS14}, however, we propose a more principled approach for learning frame-to-frame appearance and motion transitions via explicit recurrent temporal connections.

\textbf{LSTM architectures.}
LSTM networks \cite{HochreiterNC97} model sequential memories both in the long and in the short term, which makes them relevant for various sequential tasks~\cite{NgCVPR15, DonahueCVPR15, KarpathyCVPR15, JiaICCV15}.
Where early adopters used traditional features as LSTM input \cite{Baccouche2010,Baccouche2011}, more recently both Ng \etal \cite{NgCVPR15} and Donahue \etal \cite{DonahueCVPR15} propose LSTMs that explicitly model short snippets of ConvNet activations. Ng \etal demonstrate that an average fusion of LSTMs with appearance and flow input improves over improved dense trajectories \cite{wang2013iccv} and the two-stream approach of~\cite{SimonyanNIPS14}, be it that they pre-train their architecture on 1 million sports videos. Srivastava \etal \cite{SrivastavaICML15} also pre-train on hundreds of hours of sports video, but without using the video labels. Their representation demonstrates competitive results on the challenging UCF101 and HMDB51 datasets. We also rely on an LSTM architecture that combines appearance and flow for action recognition, but without the need for video pre-training to be competitive.

\textbf{ALSTM architectures.} 
Where the traditional LSTMs for action recognition emphasize on modeling the temporal extent of a sequence with the use of spatial ConvNets, Attention-LSTMs (ALSTMs) also take into account spatial locality in the form of attention. While originally proposed for machine translation \cite{BahdanauICLR15}, it was quickly recognized that a soft-Attention mechanism instead of a fixed-length vector is beneficial for vision problems as well \cite{XuICML15}. The attention turns the focus of the LSTM to particular image locations, such that the predictive capacity of the network is maximized. Sharma \etal \cite{SharmaICLR16} proposed the first ALSTM for action recognition, which proved to be an effective choice. However, by staying close to the soft-Attention architecture for image captioning by Xu \etal \cite{XuICML15}, they completely ignore the motion inside a video. Moreover, rather than vectorizing an image, for vision it is more beneficial to rely on convolutional structures \cite{lecun1998gradient,shi2015nips}. We add convolutions and motion to the ALSTM, which is not only important for the action classification, but also results in better attention for action localization. 

In addition to classifying, our model can also localize actions. 
Our approach for localization should not be confused with the recent action proposal methods, \eg \cite{JainCVPR14,YuCVPR15,GemertBMVC15,WeinzaepfelICCV15}, which generate spatio-temporal tubes, encode each of them separately and then learn to select the best one with the aid of hard to obtain spatio-temporal annotations. We propose an end-to-end video representation for action recognition that learns from just the action class label, while still being able to exploit and predict the most salient action location. We will detail our VideoLSTM architecture next.

%
\section{VideoLSTM for action classification}
VideoLSTM starts from the soft-Attention LSTM model of~\cite{XuICML15} (or ALSTM) and introduces two novel modules, the \emph{Convolutional ALSTM} and the \emph{Motion-based Attention} networks.

\noindent\textbf{Notation and terminology.}
We denote 1-d vector variables with lowercase letters, while 2-d matrix or 3-d tensor variables are denoted with uppercase letters. Unless stated otherwise, all activation functions ($\sigma(\cdot), \tanh(\cdot)$) are applied on an element-wise manner and $\odot$ is an element-wise multiplication. When implementing a particular architecture containing multiple copies of the network, \eg in unrolled networks, we refer to each of the networks copies as unit. For instance, the unrolled version of an LSTM network, see Fig.~\ref{fig:structure}, is composed of several LSTM units serially connected.

We start with a video composed of a sequence of $T$ frames and obtain the image representation $X_{1:T}=\{X_1,X_2,...,X_T\}$ for each frame using a ConvNet~\cite{krizhevsky2012imagenet,szegedy2015going,SimonyanVeryDeep}. 
Unlike previous studies~\cite{DonahueCVPR15, NgCVPR15} that use features from the last fully connected layer, we choose the convolutional feature maps, which retains spatial information of the frames. Therefore, a feature map $X_t$ at each timestep $t$ has a dimension of $N \times N \times D$, where $N \times N$ is the number of regions in an image and $D$ is the dimension of the feature vector for each region.

\subsection{Convolutional ALSTM} \label{sec:conv-lstm}

A video naturally has spatial and temporal components.
However, standard LSTM and ALSTM networks make use of full connections and treat the input as linear sequences by vectorizing the feature maps or using the fully connected activations 
from a deep convolutional network.
This results in a major drawback for handling spatio-temporal data like videos, 
since no spatial information is encoded. In order to preserve the spatial structure of the frames over time, 
we propose to replace the fully connected multiplicative operations in an LSTM unit with convolutional operations, formally
\begin{align}
I_t &= \sigma (W_{xi}*\widetilde{X}_t + W_{hi}*H_{t-1} + b_i) \\
F_t &= \sigma (W_{xf}*\widetilde{X}_t + W_{hf}*H_{t-1} + b_f) \\
O_t &= \sigma (W_{xo}*\widetilde{X}_t + W_{ho}*H_{t-1} + b_o) \\
G_t &= \tanh(W_{xc}*\widetilde{X}_t + W_{hc}*H_{t-1} + b_c) \\
C_t &= F_t \odot C_{t-1} + I_t \odot G_t  \\
H_t &= O_t \odot \tanh(C_t),
\label{eq:conv-lstm}
\end{align}
where $*$ represents the convolutional operation, $W_{x\sim}, W_{h\sim}$ are 2-d convolutional kernels. 
The gates $I_t, F_t, O_t$, the candidate memory $G_t$, memory cell $C_t, C_{t-1}$, and hidden state $H_t, H_{t-1}$ are 
3-d tensors and retain spatial dimensions as well.

\begin{figure}[t!]
    \centering 
    \includegraphics[width=0.95\linewidth]{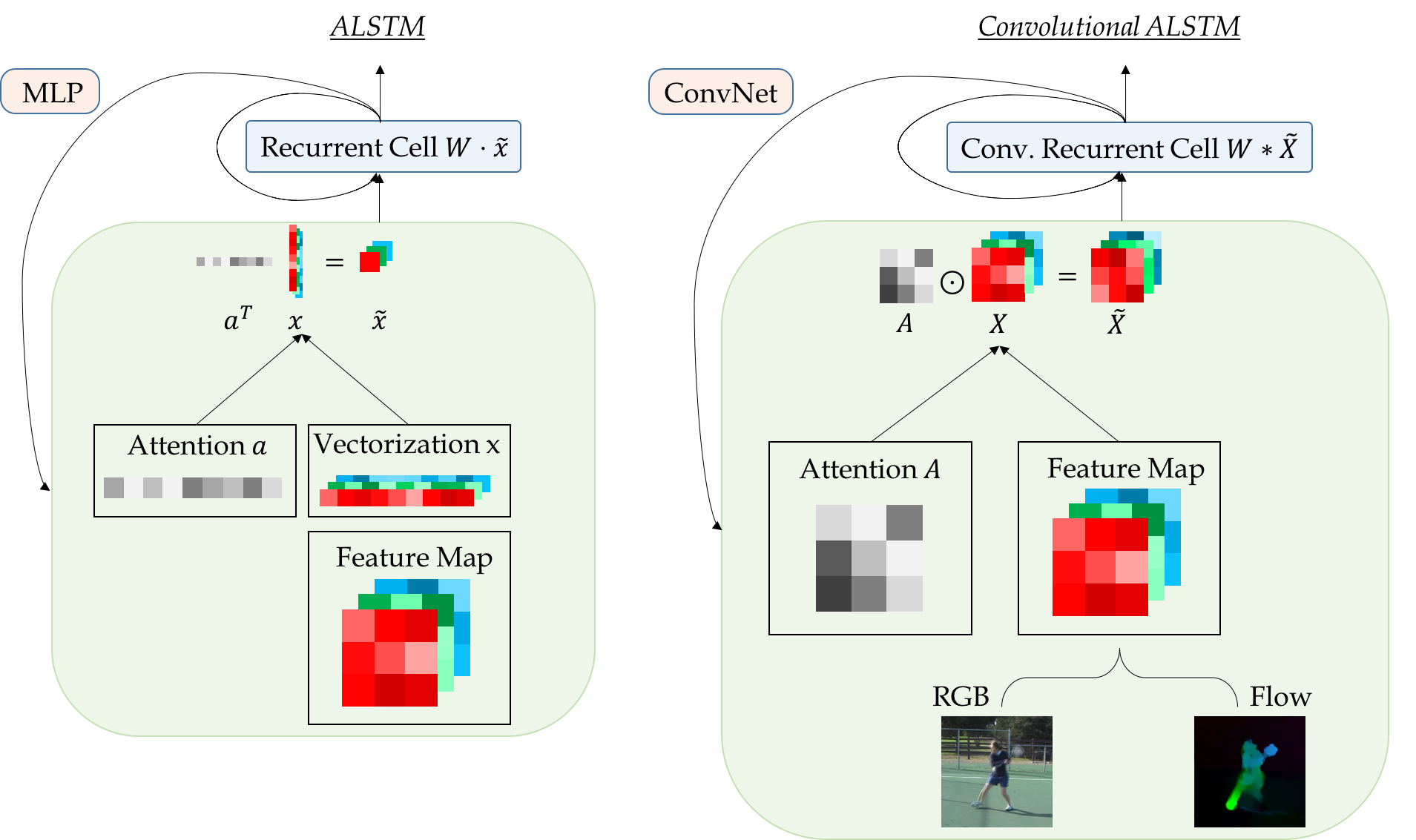}
\caption{To the left is an ALSTM model, which weighs the input vector dimensions based on attention and outputs a $D$-dimensional vector. To the right the proposed Convolutional ALSTM network, which acknowledges the two-dimensional, spatial input, performs a convolution operation and returns a $N \times N \times D$-dimensional tensor preserving the spatial structure.}
    \label{fig:convolutional_attention_lstm}
\end{figure}

Different from LSTM and ALSTM who rely on a multi-layer perceptron to generate the attention weights, we employ a shallow ConvNet with no fully connected layers conditioned on the previous hidden state and the current feature map.
More specifically, the attention map is generated by convolving the previous hidden state $H_{t-1}$ and the current input feature map $X_t$,
%
\begin{align}
Z_{t} = W_z * \tanh (W_{xa}*X_{t} + W_{ha}*H_{t-1} + b_a).
\label{eq:conv-attention}
\end{align}
By replacing the inner products with convolutions, $Z_t$ is a 2-d score map now. 
From $Z_t$ we can compute the normalized spatial attention map
\begin{align}
A_{t}^{ij} & = p(att_{ij}|X_t, H_{t-1}) = \frac{\exp(Z_{t}^{ij})}{\sum_{i}\sum_{j} \exp(Z_{t}^{ij})},
\label{eq:conv-attention-layer}
\end{align}
where $A_t^{ij}$ is the element of the attention map at position $(i,j)$.
Instead of taking the expectation over the features from different spatial locations as in ALSTM ($\tilde{x}_t = \sum_{i=1}^{N^2} a_{t}^{i} x_{t}^{i}$), we now preserve the spatial structure by weighting the feature map locations only without taking the expectation.
Formally, this is simply equivalent to an element-wise product between each channel of the feature map and the attention map
\begin{align}
\begin{split}
\widetilde{X}_{t} &= A_{t} \odot X_{t}.
\end{split}
\label{eq:conv-attend}
\end{align}
Effectively the attention map suppresses the activations from the spatial locations that have lower attention saliency.

Between the ALSTM and the Convolutional ALSTM models we spot three differences, see Fig.~\ref{fig:convolutional_attention_lstm} for an illustration.
First, by replacing the inner products all the state variables, $I_t, F_t, O_t, G_t, C_t, H_t$, of the Convolutional ALSTM model retain a spatial, 2-d structure.
As such, the video is now reckoned as a spatio-temporal medium and the model can hopefully capture better the fine idiosynchracies that characterize particular actions.
As an interesting sidenote, given that the state variables are now 2-d, we could in principle visualize them as images, which would add to the understanding of the internal workings of the LSTM units.

Second, Convolutional ALSTM resembles essentially a deep ConvNet, whose layers have recurrent connections to themselves.

Last, we should emphasize that the Convolutional ALSTM architecture can receive as input (green arrows in Fig.~\ref{fig:structure} and~\ref{fig:convolutional_attention_lstm}) and process any data with a spatial nature.
In this work we experiment with RGB and flow frames.

In the unrolled version of Fig.~\ref{fig:structure} the Convolutional ALSTM corresponds  to the upper row of LSTM units.

\subsection{Motion-based Attention}
\label{sec:motion-based-attention}

In the Convolutional ALSTM network the attention is generated based on the hidden state of the previous ALSTM unit. 
The regions of interest in a video, however, are highly correlated to the frame locations where significant motion is observed. 
This is especially relevant when attention driven recurrent networks are considered.
It is reasonable, therefore, to use motion information to help infer the attention in the ALSTM and Convolutional ALSTM network.
Specifically, we propose to add another layer with bottom-up connection with the Convolutional ASLTM layer.
This layer corresponds to the bottom row of Convolutional LSTM units of Fig.~\ref{fig:structure} in the green container, which is updated as follows
\begin{align}
I^m_t &= \sigma (W^m_{xi}*{M}_t + W^m_{hi}*H^m_{t-1} + W^m_{ei}*H_{t-1} + b^m_i) \\
F^m_t &= \sigma (W^m_{xf}*{M}_t + W^m_{hf}*H^m_{t-1} + W^m_{ef}*H_{t-1} + b^m_f) \\
O^m_t &= \sigma (W^m_{xo}*{M}_t + W^m_{ho}*H^m_{t-1} + W^m_{eo}*H_{t-1} + b^m_o) \\
G^m_t &= \tanh  (W^m_{xc}*{M}_t + W^m_{hc}*H^m_{t-1} + W^m_{ec}*H_{t-1} + b^m_c) \\
C^m_t &= F^m_t \odot C^m_{t-1} + I^m_t \odot G^m_t \\
H^m_t &= O^m_t \odot \tanh(C^m_t),
\label{eq:conv-lstm-motion}
\end{align}
where the previous hidden state from top layer $H_{t-1}$ is also given as input to the bottom layer, and $M_t$ is the feature map extracted from optical flow image at timestep $t$. 
Based on the updated LSTM cell the attention map is now conditioned on the current hidden state $H_{t}^m$ from bottom layer. This contrasts to Eq.~\eqref{eq:conv-attention-layer}, and standard LSTM architectures also, where the attention is conditioned on the $H_{t-1}$ from top layer. Namely, with the updated LSTM cell the attention at frame $t$ depends on the hidden state from the same frame $t$, instead of the previous frame $t-1$.

Moreover, note that the motion-based attention map is applied on the input feature map $X_t$ from the top layer.
Therefore, the bottom layer (green container in Fig.~\ref{fig:structure}) only helps to generate the motion-based attention and does not provide any direct information to the top layer for the final classification.

\subsubsection{VideoLSTM}
\label{sec:motion-conv-lstm}
As a last step, we define our VideoLSTM as the two layer motion-based Convolutional ALSTM architecture in Fig.~\ref{fig:structure}.

\begin{figure}[t!]
    \centering 
    \includegraphics[width=\linewidth]{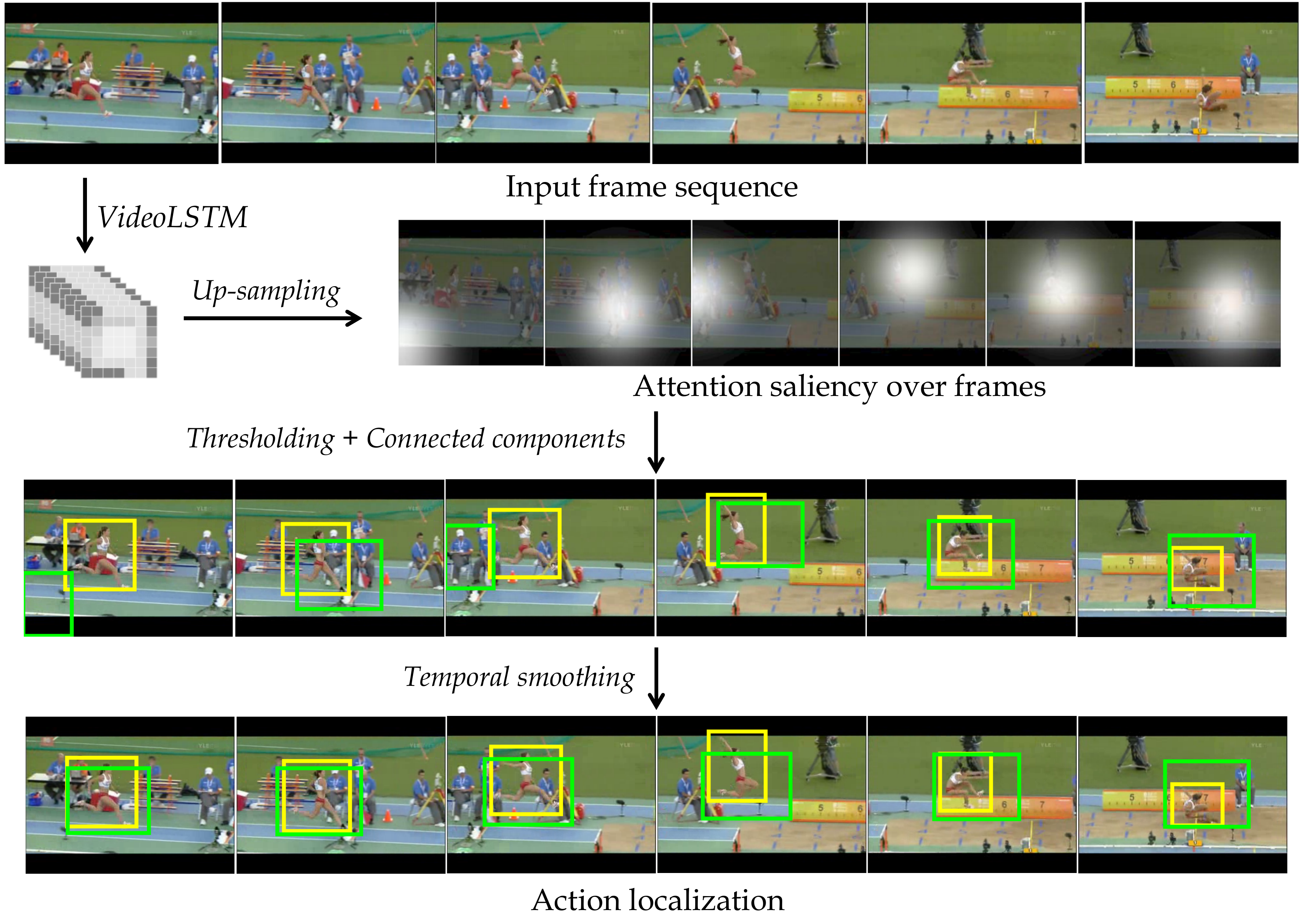}
    \caption{VideoLSTM generates attention feature maps for an input sequence of frames. These are then first up-sampled and smoothed with a Gaussian filter into saliency maps, shown superimposed over frames. Attention saliency maps are then thresholded and processed to localize the action (green box). With local temporal smoothing the boxes are temporally better aligned and lead to better localization of action compared to the ground truth (yellow box).}
    \label{fig:gen-attention-tubes}
\end{figure}

\section{VideoLSTM for action localization} \label{sec:localization}

An interesting by-product of the VideoLSTM is the sequence of attention maps, $A_t$ (Eq.~\ref{eq:conv-attention-layer}), which effectively represent the appearance and motion saliency at each frame. Each of these attention maps are first up-scaled with 2-dimensional interpolation assuming affine transform and then smoothed with a Gaussian filter, resulting into saliency maps, $\{S_1, S_2, \dots, S_T\}$ for the $T$ video frames. See Fig.~\ref{fig:gen-attention-tubes} for an example input video frame sequence. Given saliency maps $\{S_t\}$, our goal is to sample promising spatio-temporal tubes or sequences of spatial regions from the video, which are likely to bound the performed action spatially and temporally. Formally, $p^{th}$ action proposal is given by $\alpha_p=\{\alpha_p^t\}$, where $\alpha_p^t$ is a bounding-box enclosing a region from the $t^{th}$ frame.

One can think of a variety of complex models exploiting the saliency maps to formulate this problem. We take a naive, greedy approach that selects regions in frame $t$ by simply applying a threshold on saliency map $S_t$. Each connected component thus obtained forms a region, leading to a set of candidate boxes $B_t$ in $S_t$,
\begin{align}
\begin{split}
 B_t = CC_{box}(S_t \geq \theta_t)
 \end{split}
\label{eq:box_t}
\end{align}
where $CC_{box}()$ is a function generating 8-connected components enclosed by rectangular boxes and $\theta_t$ is threshold for $S_t$.

With $|B_t|$ number of boxes for frame $t$, there are $\prod_t |B_t|$ possible action proposals. Here, it is possible to generate a large number of action proposals and encode them with state-of-the-art motion features similar to~\cite{JainCVPR14,GemertBMVC15,WeinzaepfelICCV15}. However, we choose to avoid using bounding-box ground-truth and keep our approach end-to-end by generating only \emph{a single detection}.
In practice, we set high enough threshold such that we rarely get more than one box in a frame. In case of multiple boxes we select the one that has better IoU (intersection over union) with the preceding selected box. Thus we output only one proposal per video, hence dropping the subscript $p$, we have:
\begin{align}
\begin{split}
\alpha^t =   \underset{b \in B_t}{\operatorname{argmax}}  \frac{b \cap \alpha^{t-1}}{b \cup \alpha^{t-1}}
\end{split}
\label{eq:region_t}
\end{align}
Sometimes, $\alpha^t$, are sampled from the background, away from the action or actor. This is expected as the attentions from VideoLSTM is directed to focus on the locations that are discriminative for classification and context information for certain actions can be helpful. Continuing the \textit{Long Jump} example from Fig.~\ref{fig:gen-attention-tubes}, the running track distinguishes it from several other actions. Yet, most of the attention do fall on the action/actor and hence we apply temporal smoothing on the sequence of boxes, $\alpha^t$, so that sudden deviations from the action can be minimized. We use locally weighted linear regression (first degree polynomial model) to smooth the boxes, such that they do not jitter much from frame to frame. 

Other than the attention, VideoLSTM also provides classification scores for each class and each frame. We average these scores over frames to obtain confidence scores of the single proposal for each action class of interest and hence perform action localization. Two key features of our approach that most of the existing action detectors do not have are: 1) it does not require bounding box ground-truth for training and 2) it does not need to see the whole video at once, frames are processed as they are received.

\newcommand{\MCA}{MCA\xspace}
\newcommand{\ACS}{MSO\xspace}
\newcommand{\mytable}[1]{%
  \renewcommand{\arraystretch}{1.3}
  \centering
  \setlength{\tabcolsep}{8pt}
  \scalebox{0.95}{#1}
  \vspace{2mm}
}
\section{Experiments}
\subsection{Datasets}
\label{sec:dataset}
We consider three datasets for our experiments.

\textbf{UCF101~\cite{ucf101}.} This dataset is composed of about 13,000 realistic user-uploaded video clips and 101 action classes. The database is particularly interesting, because it comprises of several aspects of actions in video such as various types of activities, camera motion, cluttered background and objects/context. 
It also provides relatively large number of samples that is needed for training ConvNet/LSTM networks and hence has been popular among approaches based on deep learning. There are 3 splits for training and testing, following other recent works on end-to-end learning, \eg \cite{SimonyanNIPS14,DonahueCVPR15} we report results on the first split. Classification accuracy is used as evaluation measure.

\textbf{HMDB51~\cite{hmdb51}.} Composed of 6,766 video clips from various sources, HMDB51 dataset has 51 action classes. The dataset has two versions, the original and the one with motion stabilization, we use the more challenging original one. It has 3 train/test splits each with 3,570 training and 1,530 test videos. Following the common practice in end-to-end learning we evaluate with classification accuracy averaged over the 3 splits.

\textbf{THUMOS13 localization~\cite{THUMOS13}.} It is a subset of UCF101 with 24 classes (3,207 videos) and is provided with bounding-box level grountruth for action localization.  The dataset is quite challenging and is currently one of the largest datasets for action localization that has a rich variety of actions. The localization set consists of mostly trimmed videos with the actor mostly visible along with a few untrimmed videos also. Following the previous works~\cite{YuCVPR15,GemertBMVC15,WeinzaepfelICCV15}, we use the first split and report mean average precision (mAP) over all 24 classes.

\subsection{Implementation details} \label{sec:implementation}

\subsubsection{ConvNet architectures.} We choose the VGG-16 \cite{SimonyanVeryDeep} architecture which consists of 13 convolutional layers to train the appearance network and optical flow network. For both networks, we choose the pre-trained ImageNet model as initialization for training. 
The input of the optical flow network is a single optical flow image which has two channels 
by stacking the horizontal and vertical flow fields. 
The optical flow is computed from every two adjacent frames using the algorithm of \cite{TVL1}. We discretize the values of flow fields by linearly rescaling them to $[0, 255]$ range. We then extract the convolutional features from the last fully connected layer (\ie $fc7$) or last pooling layer (\ie $pool5$) of each VGG-16 network. Those features are fed into our LSTM architectures. 

\subsubsection{LSTM, ALSTM and VideoLSTM architectures.} 
%
%
All our LSTM models have a single layer with 512 hidden units with input feature vectors from $fc7$. For ALSTM and VideoLSTM, we use the convolutional feature maps extracted from the $pool5$ layer with size $7 \times 7 \times 512$. 
Convolutional kernels for input-to-state and state-to-state transitions are of size $3 \times 3$, while 
$1 \times 1$ convolutions are used to generate the attention map in \eq{conv-attention}. All these convolutional kernels have 512 channels. The hidden representations from the LSTM, ALSTM and VideoLSTM layer are then fully connected to an output layer which has 1,024 units. A dropout~\cite{NitishDropout} is also applied on the output before fed to the final softmax classifier with a ratio of 0.7.

For network training, we randomly sample a batch of 128 videos from the training set at each iteration. From each video, a snippet of 30 frames is randomly selected. We do not perform any data augmentation while being aware it will improve our results further. We train all our models by minimizing the cross-entropy loss using back propagation through time and $rmsprop$ \cite{RMSProp} with a learning rate of $0.001$ and a decay rate $0.9$. At test time, we follow \cite{SimonyanNIPS14} to sample 25 equally spaced segments from each video with size of 30 frames. To obtain the final video-level prediction, we first sum the LSTM frame-level class predictions over time and then average the scores across the sampled segments. For HMDB51 dataset, as the number of videos for training is very small, we use our pre-trained model on UCF101 to initialize its model. All the models are trained on an NVIDIA GeForce GTX Titan.

\subsection{Action classification}
\label{sec:result1}

\begin{table}[t!]
\renewcommand{\arraystretch}{1.3}
\centering
\scalebox{0.93}{
\begin{tabular}{l | l c c c c c}
\toprule
		& &  \textbf{ConvNet}  & \textbf{LSTM}	& \textbf{ALSTM} & \textbf{ConvLSTM} & \textbf{ConvALSTM}\\
\midrule
\multirow{2}{*}{\textbf{UCF101}} & RGB Appearance &	  77.4	  &	  77.5	  & 77.0  &	  77.6	  & 79.6 \\ 
                                 & Optical flow	  &	  75.2	  &	  78.3	  & 79.5  &	  80.4	  & 82.1 \\\hline   
\multirow{2}{*}{\textbf{HMDB51}} & RGB Appearance &   42.2    &   41.3    & 40.9  &   41.8    & 43.3 \\ 
                                 & Optical flow   &   41.8    &   46.0    & 49.2  &   48.2    & 52.6 \\     
\bottomrule
\end{tabular}}
\caption{\textbf{Convolutional ALSTM networks}. For both appearance and optical flow input, our proposed ConvALSTM improves accuracy the most.}
\label{tab:convolutional_attention_lstm}
\end{table}

\indent\textbf{Convolutional ALSTM.}
In the first experiment we compare the Convolutional ALSTM (ConvALSTM) to other architectures using appearance and optical flow input frames. For fair assessment we compare all architectures based on similar designs, implementations and training regimes. We present the results on UCF101 and HMDB51 first split in Table~\ref{tab:convolutional_attention_lstm}.

We first focus on the appearance frames. A ConvNet~\cite{SimonyanVeryDeep} already brings decent action classification, but inserting a standard LSTM on top, as suggested in~\cite{NgCVPR15,DonahueCVPR15} brings no significant benefit. The reason is that even a single RGB frame can be quite representative of an action. Moreover, since subsequent frames are quite similar, the LSTM memory adds little to the prediction. Adding soft attention~\cite{XuICML15} to the LSTM to arrive at the ALSTM proposed by~\cite{SharmaICLR16}, even deteriorates the action classification performance, while a Convolutional LSTM has a marginal impact as well. However, when considering the proposed Convolutional ALSTM on appearance data, results gain $+2.2\%$ on UCF101 and $+1.1\%$ on HMDB51. A considerable improvement, given that all other LSTM-based architectures fail to bring any benefit over a standard ConvNet.

Before discussing the impact of flow, we first note that both the original ALSTM~\cite{SharmaICLR16} and ConvLSTM models \cite{shi2015nips} rely on appearance only. A design choice which impacts their action classification potential. With flow frames all the LSTM-based models improve over the ConvNet baseline. We attribute this to the observation that in flow frames the background is not as descriptive and one can better rely on the succession of flows to recognize an action. When employing attention or convolutions independently with an LSTM we obtain a noticeable improvement over the standard LSTM, indicating that more refined LSTM architectures can exploit the flow information better. Once again, when considering the proposed Convolutional ALSTM we observe the most consistent improvement: $+3.8\%$ over LSTM, $+2.6\%$ over ALSTM and $+1.7\%$ over Convolutional LSTM on UCF101. The improvement is even larger on HMDB51.

\noindent\textbf{Motion-based Attention.} Next we evaluate the impact of motion-based attention, by explicitly modeling motion saliency using optical flow to help generate the attention maps while using RGB appearance input. We present the results in Table~\ref{tab:motion-based-attention} on UCF101 and HMDB51 first split. For ALSTM models motion-based attention obtains obvious improvements for classification, outperforming appearance-based attention by +1.6\% on UCF101 and +1.7\% on HMDB51, while for Convolutional ALSTMs we obtain also +1.5\% improvement on HMDB51. On UCF101 the improvement is smaller, because the background context is already a very strong indicator of the action class. We expect that for datasets where the background context is less indicative of the action class, motion-based attention will boost accuracy further.

\begin{table}[t!]
\renewcommand{\arraystretch}{1.2}
\centering
\begin{tabular}{l c c c c}
\toprule
                            &\multicolumn{2}{c}{\textbf{UCF101}} & \multicolumn{2}{c}{\textbf{HMDB51}} \\
                            & ALSTM & ConvALSTM & ALSTM & ConvALSTM \\
\midrule
Appearance-based Attention  & 77.0 & 79.6 & 40.9 & 43.3\\
Motion-based Attention      & 78.6 & 79.9 & 42.6 & 44.8\\
\bottomrule
\end{tabular}
\caption{\textbf{Motion-based Attention} further improves both the ALSTM model and ConvALSTM.}
\label{tab:motion-based-attention}
\end{table}

\begin{table}[t!]
\renewcommand{\arraystretch}{1.2}
\centering
\scalebox{0.78}{
\begin{tabular}{l | c c c c c c c c | c c}
    \toprule
    & \multicolumn{2}{c}{\textbf{Input}}    & \multicolumn{2}{c}{\textbf{ConvNet}} &    \multicolumn{2}{c}{\textbf{LSTM}} & \multicolumn{2}{c}{\textbf{Pre-Training}} & \textbf{UCF101} & \textbf{HMDB51} \\   
    &  RGB & Flow & Deep & Very deep & Plain & Attention & ImageNet & Sports1M & & \\ 
    \midrule    
Donahue \etal \cite{DonahueCVPR15}  & \checkmark & \checkmark & \checkmark & -- & \checkmark & -- & \checkmark &  -- & 82.9 & n/a \\
Ng \etal \cite{NgCVPR15}        & \checkmark & \checkmark & -- & \checkmark & \checkmark & -- & \checkmark & \checkmark & 88.3 & n/a \\
Srivastava \etal \cite{SrivastavaICML15}  & \checkmark & \checkmark & -- & \checkmark & \checkmark & -- & \checkmark & \checkmark & 84.3 & 44.0 \\
Sharma \etal \cite{SharmaICLR16} & \checkmark & -- & -- & \checkmark & -- & \checkmark & \checkmark & -- & 77.0 & 41.3 \\
    \midrule
This paper & \checkmark & \checkmark & -- & \checkmark & -- & \checkmark & \checkmark & -- & 89.2 & 56.4 \\
%
%
%
%
%
%
\bottomrule
\end{tabular}}
\caption{\textbf{State of the art comparison for LSTM-like architectures.} Results on UCF101 split 1 for \cite{NgCVPR15} are obtained from personal communication, the UCF101 results for \cite{SharmaICLR16} are based on our implementation of their ALSTM. \cite{SrivastavaICML15} report results on HMDB51 using RGB input only. Even without the need to train on more than 1 million sports videos, VideoLSTM is competitive.}
\label{tab:soa}
\end{table}

\begin{table}[t!]
\renewcommand{\arraystretch}{1.3}
\centering
\begin{tabular}{l c c}
\toprule
                            &\textbf{UCF101} & \textbf{HMDB51} \\
\midrule
iDT(FV)~\cite{wang2013iccv}             & 83.0  & 57.9 \\
iDT(FV)+VideoLSTM       & 91.5  & 63.0 \\   \midrule
iDT(FV)+Objects~\cite{JainCVPR15}           & 87.6  & 61.4 \\
iDT(FV)+Objects+VideoLSTM       & 92.2  & 64.9 \\   \midrule
iDT(SFV)+Objects~\cite{JainCVPR15}              & n/a   & 71.3 \\
iDT(SFV)+Objects+VideoLSTM      & n/a   & 72.9 \\   
\bottomrule
\end{tabular}
\caption{\textbf{Combination with approaches that use hand-crafted iDT features}. VideoLSTM is complementary to the approaches using iDT features. For iDT(FV), we use the software from~\cite{wang2013iccv} alongwith our implementation of classification pipeline. Stacked Fisher vectors (SFV) for HMDB51 are provided by the authors~\cite{pengECCV14stackedFisher}.}
\label{tab:idtfusion}
\end{table}

\noindent\textbf{Comparison with other LSTM architectures.}
We list in Table~\ref{tab:soa} the accuracies from other LSTM architectures for action classification. As designs and training regimes vary widely, direct comparisons are hard to make. We list properties of individual architectures to better assess relative merit. 
Instead of using average, we simply use product fusion on predictions from our VideoLSTM models with RGB input and optical flow input. VideoLSTM obtains the best result on both UCF101 and HMDB51. We note that the second best performing LSTM architecture by Ng \etal \cite{NgCVPR15}, the same architecture as the second column in Table~\ref{tab:convolutional_attention_lstm}, performs pre-training on 1 million sport videos, where our approach pre-trains on ImageNet only. Our implementation of the ALSTM of Sharma \etal obtains 77.0 mAP on UCF101 first split, where our VideoLSTM obtains 89.2, a notable difference due to the convolutions and motion-based attention. On HMDB51 the difference is even more pronounced. 
Other than LSTM models, the two-stream networks~\cite{SimonyanNIPS14} obtain 87.0\% on UCF101 and 59.4\% on HMDB51. However, the results are obtained using multi-task learning and SVM fusion.


\noindent\textbf{Combination with approaches using iDT features.}
We list in Table~\ref{tab:idtfusion} the accuracies from the state of the art approaches that use 
iDT features~\cite{wang2013iccv} for action classification. To explore the benefit results from a late fusion with 
hand-crafted features, we combine results from iDT features, Objects~\cite{JainCVPR15} and VideoLSTM. 
The prediction scores are fused by product with exponential weights. First of all, adding our VideoLSTM on top of iDT 
features improves the results significantly. This implies that our end-to-end representation is highly complementary to 
hand-crafted approaches. When combined with approach~\cite{JainCVPR15}, we achieve the state of the art performance 
92.2\% on UCF101 and 72.9\% on HMDB51. The idea from Wu \etal ~\cite{WuMM15} which fuses a CNN, an LSTM and their proposed Fusion Network, using both RGB and optical flow input obtains 91.3\% on UCF101.

\subsection{Action localization}
\label{sec:exp-action-localization}
\begin{figure}[t!]
    \centering
        \begin{subfigure}[b]{0.48\textwidth}
        \includegraphics[width=1\linewidth]{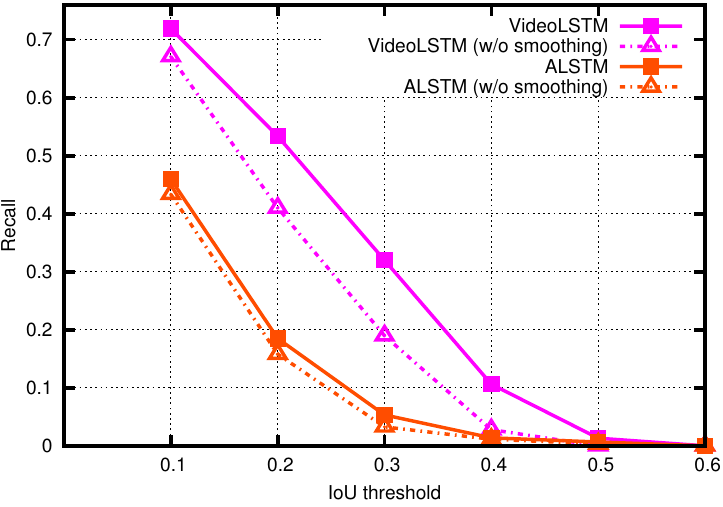}
        \caption{\small{Comparing attentions from LSTMs}}
        \label{fig:attention-tubes_lstm}
     \end{subfigure}
\hfill
    \begin{subfigure}[b]{0.48\textwidth}
        \includegraphics[width=1\linewidth]{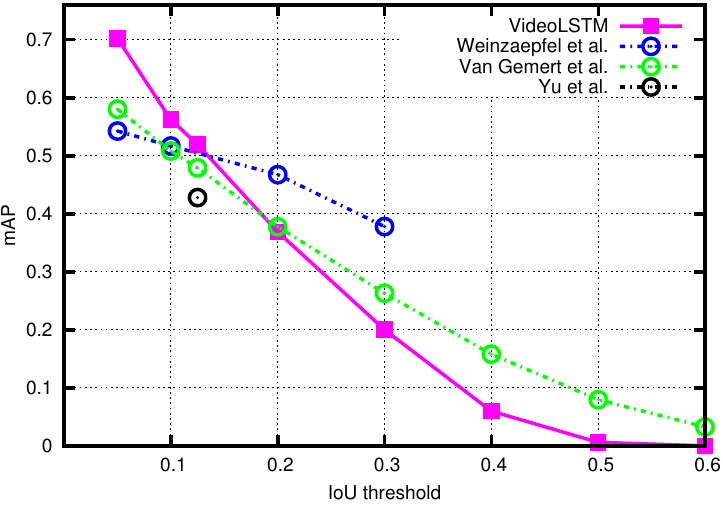}
        \caption{\small{State-of-the-art on action localization}}
        \label{fig:attention-tubes-comp}
     \end{subfigure}
        \caption{(a) VideoLSTM performs significantly better than ALSTM of Sharma \etal~\cite{SharmaICLR16}. The benefit of temporal smoothing suggests most of our attentions are on the action foreground, which does not seem to be the case for ALSTM. (b) Comparison with the state-of-the-art on action localization: our weakly supervised approach, with just one detection per video, is competitive to recent alternatives relying on bounding-box annotations during training, where our VideoLSTM only needs action class labels.}
    \label{fig:attention-tubes}
    \vspace{-5mm}
\end{figure}

In the final experiment we evaluate VideoLSTM for action localization. We set the threshold for saliency maps in Eq.~\ref{eq:box_t} to a constant ($\theta_t=100$) to ensure pixels with reasonable attention are included. We did not optimize or cross-validate over training data as it would require bounding-box ground-truth. Along with VideoLSTM, we also process the saliency maps from the ALSTM of Sharma \etal~\cite{SharmaICLR16}, and compare the two in Fig.~\ref{fig:attention-tubes_lstm}. 

There are two things to note here. First, VideoLSTM leads to strikingly higher recalls than ALSTM. Due its motion awareness and spatial structure preserving properties, VideoLSTM is capable of localizing actions, whereas ALSTM does not seem to do so.
Second, the impact of temporal smoothing is considerable in case of VideoLSTM, which means most of the attentions are on the action/actor foreground. In contrast, smoothing does not help ALSTM which suggests that the attention is either stationary or is evenly distributed between foreground and background.


In Fig.~\ref{fig:attention-tubes-comp}, we compare with the state-of-the-art methods on the THUMOS13 localization dataset with mAPs for several IoU thresholds. In contrast to VideoLSTM, all the other three approaches rely on bounding-box ground-truth to train their classifiers before applying them on action proposals. Yu \etal~\cite{YuCVPR15} only report for an IoU threshold of $0.125$. Despite using human detection their approach is about 10\% behind VideoLSTM. Weinzaepfel \etal~\cite{WeinzaepfelICCV15} use maximum supervision. In addition to bounding-box ground truth for the classification of the proposals, it also uses the bounding-box ground-truth to generate the proposals. Nonetheless, VideoLSTM does better for two of the four thresholds for which mAP is reported in~\cite{WeinzaepfelICCV15}. A considerable achievement, given that we only need an action class label. Van Gemert \etal~\cite{GemertBMVC15} uses the bounding-box ground-truth for classifying their APT action proposals. Despite using only video 
level annotation, our weakly supervised approach is competitive to APT~\cite{GemertBMVC15}, is better than Weinzaepfel \etal~\cite{WeinzaepfelICCV15} for the lower thresholds and outperforms Yu \etal~\cite{YuCVPR15}. Note that these methods also make use of thousands of spatio-temporal action proposals~\cite{GemertBMVC15} or object proposals~\cite{WeinzaepfelICCV15}. 
Several visual examples of action localization are added in the supplementary material.
We conclude that VideoLSTM returns surprisingly good \emph{hit-or-miss} action localization, 
especially considering that only a single spatio-temporal proposal is returned.

\section{Conclusion}

In this work we postulate that to model videos accurately, we must adapt the model architecture to fit the video medium and not vice versa. The video model must, therefore, address common video properties, \ie what is the right appearance and motion representation, how to transform spatio-temporal video content into a memory vector, how to model the spatio-temporal locality of an action, \emph{simultaneously} and not in isolation. We propose \textit{VideoLSTM}, a new recurrent neural network architecture intended for action recognition. VideoLSTM makes three novel contributions to address these video properties in a joint fashion: \emph{i)} it introduces convolutions to exploit the spatial correlations in images, \emph{ii)} introduces shallow convolutional neural network to allow for motion information to generate motion-based attention maps and finally \emph{iii)} by only relying on video-level action class labels exploits the attention maps to localize the action spatio-temporally. Experiments on three challenging datasets outline the theoretical as well as the practical merits of VideoLSTM.


\bibliographystyle{splncs}
\bibliography{bib}
\end{document}